\documentclass[runningheads]{llncs}

\usepackage{eccv}

\usepackage{amsmath,amsfonts,bm}

\def\eqref#1{equation~\ref{#1}}
\def\1{\bm{1}}

\def\vtheta{{\bm{\theta}}}

\DeclareMathAlphabet{\mathsfit}{\encodingdefault}{\sfdefault}{m}{sl}
\SetMathAlphabet{\mathsfit}{bold}{\encodingdefault}{\sfdefault}{bx}{n}

\newcommand{\E}{\mathbb{E}}

\newcommand{\Var}{\mathrm{Var}}

\usepackage{eccvabbrv}

\usepackage{graphicx}
\usepackage{booktabs}

\usepackage[accsupp]{axessibility}  % Improves PDF readability for those with disabilities.

\usepackage[dvipsnames]{xcolor}
\newcommand\rachel[1]{\textcolor{blue}{[RL: #1]}}
\newcommand\sv[1]{\textcolor{teal}{[SV: #1]}}
\newcommand\hy[1]{\textcolor{cyan}{[HY: #1]}}
\newcommand\mw[1]{\textcolor{red}{[MW: #1]}}
\newcommand\as[1]{\textcolor{green}{[AS: #1]}}

\renewcommand\rachel[1]{}     %uncomment to hide rachel
\renewcommand\sv[1]{}         %uncomment to hide sushant
\renewcommand\hy[1]{}        %uncomment to hide hank
\renewcommand\mw[1]{}     %uncomment to hide michael
\renewcommand\as[1]{}     %uncomment to hide apoorva

\newcommand{\name}{X4Val\xspace}
\newcommand{\method}{\mathrm{X4Val}}   % for math subscripts

\newcommand{\metric}{F}
\newcommand{\policyundertest}{\pi^*}
\newcommand{\muval}{\mu}
\newcommand{\mcmean}{\hat{\mu}_{\mathrm{MC}}}
\newcommand{\ourmean}{\hat{\mu}_{\method}}

\newcommand{\scenario}{X}
\newcommand{\unlabeledscenario}{\scenario^{\mathrm{unlab}}}
\newcommand{\targetdist}{P_X}
\newcommand{\scenariodist}{\targetdist}  % backwards-compatible alias
\newcommand{\auxsignal}{G}               % inexpensive signal / auxiliary measurement
\newcommand{\embedding}{z}
\newcommand{\embedfn}{\phi}

\newcommand{\param}{\vtheta}
\newcommand{\surrogatefn}{f_\param}
\newcommand{\surrogatepred}{\widehat{\metric}}

\newcommand{\mcf}{\surrogatefn}

\newcommand{\labeleddata}{\mathcal{D}_{\mathrm{lab}}}
\newcommand{\traindata}{\mathcal{D}_{\mathrm{train}}}
\newcommand{\estdata}{\mathcal{D}_{\mathrm{est}}}
\newcommand{\unlabeleddata}{\mathcal{D}_{X}}

\newcommand{\auxdata}{\mathcal{D}^{\mathrm{aux}}}

\newcommand{\numreal}{n}

\newcommand{\numest}{n_{\mathrm{est}}}
\newcommand{\numunlabeled}{m}

\newcommand{\numtasks}{J}

\newcommand{\numfolds}{K}
\newcommand{\foldidx}{k}

\usepackage{hyperref}

\usepackage{orcidlink}

\begin{document}

\title{X4Val: Learning Neural Surrogates for Variance-Reduced Policy Evaluation}

\titlerunning{X4Val}

\author{Rachel Luo\inst{1} \and 
Michael Watson\inst{1} \and
Apoorva Sharma\inst{1} \and
Heng Yang\inst{1,2} \and
Han Qi\inst{2} \and
Edward Schmerling\inst{1} \and
Sushant Veer\inst{1} \and
Boris Ivanovic\inst{1} \and
Marco Pavone\inst{1,3}}

\authorrunning{R. Luo et al.}
\institute{NVIDIA Research, Santa Clara, CA, USA \\
\email{\{raluo, mwatson, apoorvas, eschmerling, hengy, sveer, bivanovic, mpavone\}@nvidia.com} \and
Harvard University, Cambridge, MA, USA \\
\email{\{hqi\}@g.harvard.edu} \and
Stanford University, Stanford, CA, USA }

\maketitle

\begin{abstract}

Rigorous evaluation of learning-based robotic systems is an essential prerequisite for deployment. However, real-world test data is expensive to gather; moreover, in a typical iterative development context, data gathered from the latest policy is necessarily limited in scale. This motivates evaluation methodologies that make use of heterogeneous data sources, including simulation, historical policy logs, and data collected from related platforms or environments. While such auxiliary data are abundant and inexpensive, they are generally not directly representative of real-world outcomes---for example, performance in simulation may differ substantially from performance in the real world---making their principled use for high-confidence performance estimation challenging. In this paper, we introduce \name, a general framework for variance-reduced real-world metric estimation in the presence of non-paired, multi-domain data. \name embeds samples from real and auxiliary domains into a shared representation space and learns a transferable predictor of real-world metrics; this learned predictor is then incorporated into a control-variates estimator, enabling variance reduction even when paired samples are unavailable. We provide theoretical analysis and empirical evaluations on autonomous driving and real-world robot manipulation tasks, domains across which \name achieves up to 38.4\% variance reduction and demonstrates consistent improvements over strong baselines. These results show that non-paired, heterogeneous data can be leveraged to substantially improve the sample efficiency of rigorous robotic system validation.

\end{abstract}

\section{Introduction}
\label{sec:intro}

\begin{figure*}[t]
  \centering
  \includegraphics[width=0.9\textwidth]{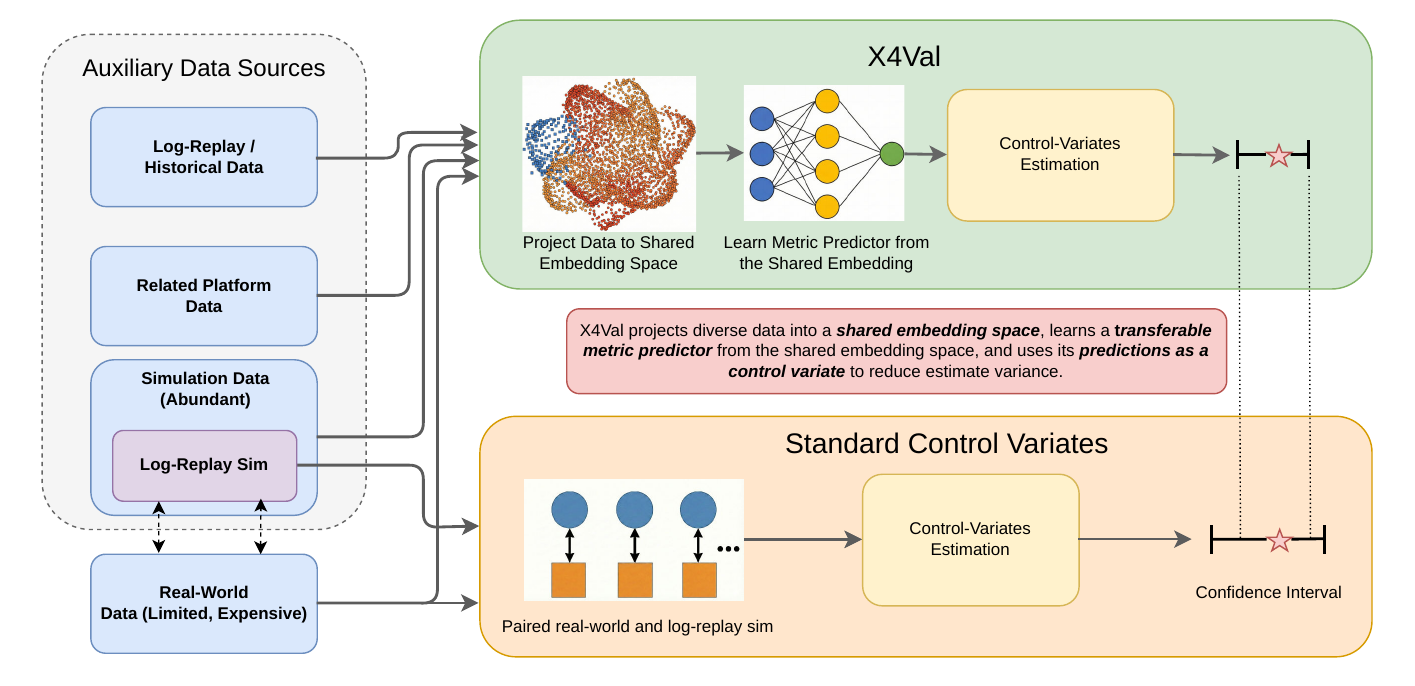}
  \caption{\footnotesize Comparison of \name with standard control variates-based estimation. \textbf{Standard approaches (bottom)} can only use real world data with its log-replay simulations enforcing a strict pairing between the real and simulated data and limiting the amount of data that can be used. \textbf{\name (top)} on the other hand, can use a diverse range of data sources by projecting them to a shared embedding space and learning a transferrable metrics predictor from this space. The ability to leverage significantly more data helps in reducing the variance of the metric estimates.}
  \label{fig:x4val-anchor}
\end{figure*}

Rigorous evaluation and validation of robotic systems is essential for both scientific progress and real-world deployment, particularly in safety-critical domains such as autonomous driving, mobile manipulation, and human–robot interaction. Such validation requires accurate and high-confidence estimates of performance metrics under the distribution of conditions expected at deployment time. However, obtaining sufficient real-world data to support such estimates is often infeasible: real-world testing is expensive, time-consuming, and may expose systems to safety risks, making large-scale data collection impractical in many settings.

Yet in practice, robotic systems are seldom developed in isolation. Practitioners often have access to abundant auxiliary data, including simulation rollouts, historical logs from earlier policy versions, evaluations from related platforms, or data collected in different environments.
Such ambient auxiliary data are typically far cheaper to acquire and more abundant than real-world on-policy measurements; the question is how to access the valuable information they contain about system behavior given that these data sources generally differ from the true deployment setting and cannot be treated as direct substitutes for real-world evaluation. 

Recent variance-reduction approaches for autonomous system validation have shown that auxiliary measurements can substantially reduce real-world sample requirements when they are available in paired form~\cite{luo2025_sim2val, badithela2025reliable}. In particular, control-variate estimators can exploit correlations between surrogate and real-world measurements (e.g. simulation and real-world rollouts) collected on the same scenarios or initial conditions. However, this reliance on sample-wise pairing limits their applicability. In many realistic settings, there is no applicable notion of pairing.
For example, real-world data may have been collected under earlier policy versions, while surrogate evaluations are available only for newer policies; simulation data may be generated on different inputs or environments than those observed in the real world; or data may be collected across multiple platforms with no direct correspondence between samples. 
Thus, in these regimes, large quantities of potentially useful auxiliary data are available, but classical control-variate methods cannot directly exploit them.

In this paper, we introduce \name, a general framework for variance-reduced real-world metric estimation in non-paired, multi-domain settings. The central idea of \name is to replace or augment explicit sample-wise pairing with a \emph{learned neural surrogate} of the real-world metric, trained by leveraging abundant auxiliary data collected across multiple platforms, environments, or policy versions. Rather than requiring a one-to-one correspondence between surrogate and real-world evaluations, 
\name learns a transferable predictor of the real-world metric that captures cross-domain structure through representation learning, transfer learning, and meta-learning. This learned neural surrogate is then used as a statistically valid control variate, enabling variance reduction even when direct pairing is absent; see Fig.~\ref{fig:x4val-anchor}. 

Concretely, \name proceeds in three stages. First, samples from all available domains are embedded into a shared representation space that captures task-relevant structure. 
Second, a neural surrogate is trained to predict real-world metrics from these embeddings and auxiliary features, while a cross-fitting procedure inspired by cross-prediction-powered inference improves sample efficiency in the low-label regime~\cite{zrnic2024cppi}.
Finally, the learned surrogate is incorporated into a control-variates estimator, enabling variance reduction even in the absence of paired samples. 

This framework generalizes classical control-variates approaches for validation, extending them to substantially broader and more realistic evaluation scenarios.
\name is particularly useful in settings where robotic systems are iteratively updated and evaluated over time, where historical data from previous policies remain informative; when policies are trained in one environment or geography and deployed in another; when validation must combine data from multiple simulators or hardware platforms; or when real-world testing is limited but large-scale surrogate evaluations are available across related domains.

We illustrate these capabilities through several case studies that mirror common robotic development pipelines, described below. Across these scenarios, \name provides a principled mechanism for exploiting heterogeneous, non-paired data without sacrificing statistical rigor.

\textit{Geographical transfer.}
When deploying a policy trained in one geographical region to another, validation data from the original region remains informative but does not match the new deployment distribution. 
In this setting, \name uses source-region evaluations as auxiliary data to reduce uncertainty in target-region performance estimates, improving over both naive Monte Carlo and region-specific control-variate estimators.

\textit{Sequential policy development.}
In large-scale autonomy programs, policies are iteratively trained on progressively larger datasets and evaluated at each stage. 
In this setting, \name leverages historical evaluations across policy versions to produce tighter confidence intervals for the current policy than methods that use only current paired samples.

\textit{Cross-platform evaluation.}
Robotic systems are often developed and evaluated across multiple platforms, such as simulation environments and physical hardware. While large-scale evaluations may be readily available in simulation or other surrogate platforms, real-world evaluation is typically limited due to cost, time, and safety constraints. Moreover, evaluations across platforms are rarely paired: the exact initial states or scenarios observed on the physical system may not be reproducible in simulation, and vice versa. 
\name exploits this non-paired surrogate data to reduce uncertainty in real-world metric estimates, demonstrating variance reduction even when classical paired control variates are not applicable.

Our contributions are as follows:
\begin{enumerate}
    \item We introduce \name, a framework for variance-reduced estimation of real-world performance metrics in robotic validation settings using heterogeneous, non-paired auxiliary data sources.
    \item We develop a learning-based control-variate estimator that uses transferable neural surrogates trained from limited real-world labels and abundant auxiliary-domain data. 
    \item We show that the resulting estimator preserves statistical validity and generalizes classical paired control-variate approaches for validation, retaining them as a special case while extending variance reduction to settings without sample-wise pairing.
    \item Through several case studies reflecting common robotics development pipelines--including geographical transfer, sequential policy development, and cross-platform evaluation--we demonstrate that \name substantially reduces estimator variance and yields tighter confidence intervals compared to standard Monte Carlo and classical control-variate baselines.
\end{enumerate}

\section{Problem Setup}

Formally, our goal is to estimate the expected value of a metric of interest $\metric$ which measures the performance of a policy under test $\policyundertest$ over a target scenario distribution $\scenariodist$, i.e., $\E_{\scenario\sim \scenariodist}[\metric(\scenario)]$. 
The standard approach to estimating this expectation is to use the Monte-Carlo mean estimator, taking \textit{on-policy} measurements of $\metric$ by directly testing $\policyundertest$ on a representative (i.e., assumed to be i.i.d.) set of sample scenarios $\scenario$ drawn from the scenario distribution $\scenariodist$, 
\begin{equation*}
    \mcmean = \frac{1}{\numreal} \sum_{i=1}^\numreal \metric_i,
\end{equation*}
where $\numreal$ is the number of measurements collected, $\metric_i \sim \Pr(\metric \mid \scenario_i)$, and $\scenario_i \sim \scenariodist$. 
This estimator is unbiased and has a variance inversely proportional to the number of samples collected:
\begin{align*}
    \Var(\mcmean) = \frac{1}{\numreal} \Var(\metric).  
\end{align*} 

In this work, we explore how auxiliary data can improve the efficiency of this validation task. We distinguish between two types of additional data. First, we may have access to target-domain scenario-only data 
\[ \unlabeleddata = \{(\unlabeledscenario_i,\auxsignal_i)\}_{i=1}^{\numunlabeled}, 
\qquad \unlabeledscenario_i \sim \targetdist, \] 
for which the inexpensive inputs \((\unlabeledscenario_i,\auxsignal_i)\) are available but the expensive metric value \(\metric_i\) is not observed. Such data may come from logged deployment-domain scenarios, human-driven data collection, inexpensive setup states, or other sources that sample the target deployment distribution. 
Second, we may have access to auxiliary datasets 
\[ \auxdata_1,\ldots,\auxdata_\numtasks \] 
from related domains, such as simulations, historical policy versions, related platforms, or different deployment environments. These auxiliary datasets may contain metric measurements, surrogate measurements, features, or other signals useful for learning about system behavior. However, they need not be sampled from the target distribution \(\targetdist\), and therefore cannot in general be treated as direct substitutes for target-domain validation data.

Our goal is to construct an estimator of \(\muval\) that remains unbiased for the target deployment distribution while achieving lower variance than \(\mcmean\) for the same number of expensive target-domain metric measurements. The key idea is to use auxiliary data to learn a neural surrogate of the target metric, and then use target-distribution samples to form a control-variate estimator whose residual correction preserves unbiasedness.

\section{Related Work}

Improving the data-efficiency of metric estimation is an important research topic across many fields, especially where the costs of obtaining measurements is high. In the Monte-Carlo estimation literature, there are a variety of techniques for \textit{variance reduction}, such as importance sampling, stratified sampling, and control variates, which provide practitioners methods to leverage domain knowledge to achieve lower-variance estimates of the mean with the same number of measurements \cite{owen2013monte}. In particular, the concept of control variates -- correlated signals whose expectation is more easily estimated -- has been applied to the task of empirical validation. 
Prediction-powered inference \cite{angelopoulos2023prediction, angelopoulos2023ppi++, zrnic2024cppi} applies a similar idea to improve general statistical estimation by using a prediction model as a control variate. These ideas have been applied to the evaluation of various models, from reducing expensive human feedback costs in LLM evaluation\cite{zhou2025accelerating, boyeau2024autoeval}, to reducing  testing costs in robot policy evaluation \cite{badithela2025reliable, luo2025_sim2val}.

A central challenge in applying control variates to empirical validation is obtaining a suitable correlated signal. SureSim~\cite{badithela2025reliable} and Sim2Val~\cite{luo2025_sim2val} leverage recent advances in high-fidelity ``real-to-sim'' pipelines, where a logged real-world interaction can be reconstructed and re-tested in simulation~\cite{caesar2020nuscenes, gulino2023waymax}; the resulting simulation measurement serves as a control variate for the real-world metric. Sim2Val further applies a learned correction to the simulation measurement to improve correlation and thereby improve estimation efficiency. While effective, these approaches rely on sample-wise paired evaluations of the target system and a surrogate system on the same scenarios or initial conditions. 
\name{} builds on these ideas---using correlated surrogate signals and learned predictors to reduce validation variance---but removes the need for explicit pairing by learning a transferable neural surrogate from heterogeneous auxiliary data sources.

Our problem setting is also related to policy evaluation in reinforcement learning, particularly off-policy, off-environment, and off-domain evaluation. In offline reinforcement learning, variance-reduction techniques are often used to estimate policy performance before deployment~\cite{levine2020offline, precup2000eligibility}. These methods typically correct for distribution shift between the data-collection policy and the target policy using importance sampling, doubly robust estimation, or occupancy-ratio corrections~\cite{jiang2016doubly}. Recent off-environment and off-domain evaluation methods similarly use source-domain or offline data to reduce the cost and risk of target-domain evaluation~\cite{Katdare2023MarginalizedIS, Niu2024ACS}. Mandyam et al.~\cite{mandyam2025perry} propose PERRY, which combines auxiliary rollout data with prediction-powered inference to obtain confidence intervals for policy evaluation. These works share our high-level motivation, but they are generally formulated in the reinforcement-learning setting, where the system is described through state-action trajectories and where action likelihoods, transition models, or reliable occupancy-ratio estimates may be available. In contrast, \name{} targets a more general black-box validation setting: the system may be a full autonomy stack, the metric may be an arbitrary scenario-level outcome, and action likelihoods, transition models, or occupancy ratios may be unavailable or ill-defined. 

Finally, \name{} draws on the broader machine learning literature on transfer learning, multi-task learning, and meta-learning. Transfer learning studies how a model trained on one prediction task can be adapted to another, while multi-task and meta-learning methods use experience across related tasks to improve performance or enable rapid adaptation to new tasks. A common paradigm is pretrain-finetune, in which models are first trained on large source datasets and subsequently adapted to downstream tasks using limited labeled data; this approach is widely used across modern vision and language systems~\cite{Tan2018ASO, Jiang2022TransferabilityID, Chato2023SurveyOT, Devlin2019BERTPO, Yosinski2014HowTA}. Closely related amortized meta-learning methods learn shared inference or adaptation networks that map small task datasets to task-specific predictors or parameters~\cite{Iakovleva2020MetaLearningWS, Ravi2018AmortizedBM, Zhang2025AmortizedBM}. \name{} uses similar transfer and meta-learning mechanisms to learn a neural surrogate from auxiliary-domain data, but uses the resulting predictor specifically as a control variate for statistically valid target-domain metric estimation.

\section{Method}
\label{sec:method}
\name{} improves validation efficiency by learning a neural surrogate of the target metric and using this surrogate as a control variate. A key distinction in our setting is that heterogeneous auxiliary data may be used freely to improve the learned surrogate, but only samples drawn from the target deployment distribution are used in the estimator itself. This allows \name{} to benefit from related data sources without treating them as direct substitutes for target-domain validation data. At a high level, \name{} proceeds in three steps. First, data from the target and auxiliary domains are mapped into a shared representation space. Second, a neural surrogate is trained to predict the target metric using limited target-domain labels together with auxiliary data from related domains. Finally, this surrogate is used as a control variate to estimate the target-domain mean metric and construct confidence intervals.

\subsection{Projecting Heterogeneous Data into a Shared Embedding Space}

The data available during autonomous system development are often heterogeneous: they may come from different platforms, simulators, policy versions, sensor modalities, or deployment environments. To make these data useful for learning a shared metric predictor, \name{} first maps each scenario into a common representation space.

Let $\embedding = \embedfn(\scenario, \auxsignal)$
denote a representation of a scenario \(\scenario\) together with any inexpensive signal \(\auxsignal\) available for that scenario. Depending on the application, $\embedfn$ may include hand-designed scenario features, learned embeddings from logged sensor observations, foundation-model features, simulator-derived quantities, open-loop metrics, or other auxiliary measurements that can be computed without the expensive target-domain validation procedure. In general, $\embedfn$ should capture task-relevant structure that helps predict the metric of interest. 

\subsection{Learning a Transferable Neural Surrogate}
\label{sec:learning_a_transferable_neural_surrogate}
With all data mapped to a shared feature space, \name{} next learns a neural surrogate of the target metric,
\begin{equation}
\mcf : \embedding \mapsto \surrogatepred,
\end{equation}
where \(\surrogatefn(\embedding)\) predicts the target metric \(\metric\) from the embedded scenario representation. 

Let 
\[ \labeleddata = \{(\scenario_i,\auxsignal_i,\metric_i)\}_{i=1}^{\numreal}, \qquad \scenario_i \sim \targetdist, \] 
denote the available labeled target-domain metric data, where observing \(\metric_i\) requires the expensive target-domain validation procedure. We split this dataset into disjoint training and estimation subsets, 
\[ \labeleddata = \traindata \cup \estdata, \qquad \traindata \cap \estdata = \emptyset, \] 
where \(\traindata\) is used to adapt or train the neural surrogate, and \(\estdata\) is reserved for estimating the residual correction. Let \(\numest = |\estdata|\). 

In addition, let 
\[ \unlabeleddata = \{(\unlabeledscenario_i,\auxsignal_i)\}_{i=1}^{\numunlabeled}, \qquad \unlabeledscenario_i \sim \targetdist, \] 
denote target-domain scenario samples for which the inexpensive surrogate inputs can be computed, but for which the expensive metric \(\metric\) need not be observed. These samples may come from logged target-domain scenarios, human-driven data collection, inexpensive simulation setup states, or any other source that samples the same deployment scenario distribution. 

The neural surrogate is trained using the target training data together with auxiliary datasets from related domains, 
\begin{equation} 
\mcf = \mathcal{A} \left( \traindata, \auxdata_1,\ldots,\auxdata_\numtasks \right), 
\end{equation} 
where \(\mathcal{A}\) denotes the chosen transfer-learning or meta-learning algorithm. For example, one may pretrain the surrogate on auxiliary domains and finetune it on target-domain metric data, train a multi-task model across multiple auxiliary signals, or use an amortized meta-learning procedure that adapts the surrogate to a new target domain or policy using a small number of labeled examples.

Importantly, the auxiliary datasets \(\auxdata_1,\ldots,\auxdata_\numtasks\) do not need to be drawn from the target distribution \(\targetdist\); any auxiliary data which provides a useful learning signal that transfers to the target domain are useful. Note that the final estimator, which we detail below, remains unbiased for any surrogate $\surrogatefn$ that is chosen independently from the estimation data, which we ensure by training the surrogate on a disjoint split from the data used for estimation. 

\subsection{Variance-Reduced Estimation with a Neural Surrogate}

Given the learned surrogate \(\surrogatefn\), we estimate the target mean 
\( \muval = \E_{\scenario\sim\targetdist}[\metric] \) 
using 
\begin{equation} 
\begin{aligned}
\ourmean &= \frac{1}{\numunlabeled} 
\sum_{(\unlabeledscenario_i,\auxsignal_i)\in\unlabeleddata} 
\surrogatefn\!\left( \embedfn(\unlabeledscenario_i,\auxsignal_i) \right) \\ 
&\qquad\qquad + \frac{1}{\numest} \sum_{(\scenario_i,\auxsignal_i,\metric_i)\in\estdata} 
\left[ \metric_i - \surrogatefn\!\left( \embedfn(\scenario_i,\auxsignal_i) \right) \right].
\end{aligned}
\label{eq:x4val-single-split} 
\end{equation} 
The first term estimates the target-domain expectation of the learned neural surrogate, while the second term estimates the residual correction between the true metric and the surrogate.

For any fixed learned surrogate \(\surrogatefn\) trained without using the labels in \(\estdata\), the estimator in Eq.~\ref{eq:x4val-single-split} is unbiased provided that \(\unlabeleddata\) and \(\estdata\) are drawn from the target deployment distribution:
\begin{align*}
\E[\ourmean] &= \E_{\scenario\sim\targetdist}
\left[
    \surrogatefn(\embedfn(\scenario,\auxsignal))
\right]
+ \E_{\scenario\sim\targetdist}
\left[
    \metric - \surrogatefn(\embedfn(\scenario,\auxsignal))
\right] \\
&= \E_{\scenario\sim\targetdist}[\metric] \\
&= \muval .
\end{align*}
Thus, the neural surrogate need not be an unbiased predictor of the target metric. It may be trained using biased, shifted, or heterogeneous auxiliary data; any systematic error in the surrogate is corrected by the residual term computed on target-domain metric samples.

Assuming \(\unlabeleddata\) and \(\estdata\) are independent, the conditional variance of Eq.~\ref{eq:x4val-single-split} is 
\begin{equation} 
\begin{aligned} 
\Var(\ourmean \mid \surrogatefn) &= \frac{1}{\numunlabeled} \Var_{\scenario\sim\targetdist} \left[ \surrogatefn(\embedfn(\scenario,\auxsignal)) \right] \\ 
&\quad+ \frac{1}{\numest} \Var_{\scenario\sim\targetdist} \left[ \metric - \surrogatefn(\embedfn(\scenario,\auxsignal)) \right]. 
\end{aligned}
\label{eq:x4val-variance} 
\end{equation} 
This expression highlights the role of the learned neural surrogate. A useful surrogate should make the residual \(\metric-\surrogatefn(\embedfn(\scenario,\auxsignal))\) low variance while not introducing excessive variance in the surrogate itself. When \(\numunlabeled\) is large, the first term is small, and the primary objective is to learn a surrogate whose fluctuations are strongly correlated with those of the target metric. When \(\numunlabeled\) is limited, the variance of the surrogate over target-domain scenarios starts to have a higher impact.

This observation suggests that the learning objective for \(\surrogatefn\) should be aligned with variance reduction rather than prediction accuracy alone. For example, one may train or post-process the surrogate to minimize an empirical estimate of 
\begin{equation} 
\mathcal{L}_{\name} = \widehat{\Var}_{\estdata} \left[ \metric - \surrogatefn(\embedfn(\scenario,\auxsignal)) \right] + \frac{\numest}{\numunlabeled} \widehat{\Var}_{\unlabeleddata} \left[ \surrogatefn(\embedfn(\scenario,\auxsignal)) \right]. 
\label{eq:x4val-loss} 
\end{equation} 
This differs from a standard training objectives like mean-squared-error: the estimator is insensitive to the bias of the surrogate, but sensitive to the variance of the residual and, when \(\numunlabeled\) is of comparable magnitude to $\numest$, to the variance of the surrogate over target-domain scenarios.

Alternatively, when using an off-the-shelf transfer-learning strategy, the learned surrogate can be post-hoc scaled before being used as a control variate. Let \(\tilde f_\param\) denote a predictor obtained from pretraining, finetuning, or meta-learning, and set 
\[ \surrogatefn(\embedding) = \beta \tilde f_\param(\embedding). \] 
The scaling parameter \(\beta\) can be chosen on held-out target-domain data to minimize the empirical analogue of Eq.~\ref{eq:x4val-variance}. This recovers the standard control-variate intuition: the learned surrogate need not be perfectly calibrated, but it should provide a signal whose fluctuations are correlated with the fluctuations of the target metric.

\paragraph{Cross-fitting for improved sample efficiency.}
For clarity, we outlined the \name{} approach above for a single split of available data into $\estdata$ and $\traindata$. In practice, it is more data-efficient to apply a cross-validation style approach, similar to \cite{zrnic2024cppi}, splitting available data into $\numfolds$ disjoint folds. Treating each fold as an estimation dataset $\estdata$, we compute the estimate according to Eq.~\ref{eq:x4val-single-split}, with the neural surrogate trained on the data from the other folds. Averaging these $\numfolds$ estimators yields an unbiased estimator with lower variance, as all target domain data contributes to the estimation of the residual, at the expense of additional computation in model training. 
We use the cross-fitted estimator in our experiments, see Appendix~\ref{app:crossfit} for a more detailed exposition and details on confidence-interval construction.

\subsection{Obtaining Confidence Intervals}

The estimator in Eq.~\ref{eq:x4val-single-split} yields confidence intervals through the Central Limit Theorem. For the single-split estimator, let 
\[ \hat{\sigma}^2_f = \widehat{\Var}_{(\unlabeledscenario_i,\auxsignal_i)\in\unlabeleddata} \left[ \surrogatefn(\embedfn(\unlabeledscenario_i,\auxsignal_i)) \right] \] 
denote the empirical variance of the learned surrogate over target-domain scenario-only samples, and let 
\[ \hat{\sigma}^2_\Delta = \widehat{\Var}_{(\scenario_i,\auxsignal_i,\metric_i)\in\estdata} 
\left[ \metric_i - \surrogatefn(\embedfn(\scenario_i,\auxsignal_i)) \right] \] 
denote the empirical variance of the residual correction over held-out target-domain metric samples. Then an approximate \((1-\alpha)\) confidence interval for \(\muval\) is 
\begin{equation} 
\mathcal{C}_\alpha = \left[ \ourmean \pm z_{1-\alpha/2} \sqrt{ \frac{\hat{\sigma}^2_f}{\numunlabeled} + \frac{\hat{\sigma}^2_\Delta}{\numest} } \right]. 
\label{eq:x4val-ci} 
\end{equation}

The same form applies to the cross-fitted estimator, with \(\hat{\sigma}^2_f\) computed from the averaged fold-wise surrogate predictions on target-domain scenario-only samples, and \(\hat{\sigma}^2_\Delta\) computed from the held-out residuals across all labeled target-domain samples; see Appendix~\ref{app:crossfit}. In all cases, the confidence interval targets the deployment-domain mean \(\muval\), and its validity relies on the samples used in the estimator---the target-domain scenario-only samples and the target-domain metric samples---being drawn from \(\targetdist\). Auxiliary data from other domains can improve the learned neural surrogate, but do not themselves define the target expectation.

\section{Autonomous Driving Case Studies}

In this section, we demonstrate the efficacy of \name in improving the efficiency of performance estimation in case studies that illustrate common development processes in the autonomous vehicle industry. For these experiments, we study the performance of a lightweight motion-planning policy trained on subsets of the NVIDIA PhysicalAI-Autonomous-Vehicles dataset~\cite{nvidia2025physical_ai_av}. The policy maps a DINOv3 embedding of the front camera view and egomotion history to a planned trajectory, and is evaluated in both open-loop and closed-loop using the AlpaSim simulator~\cite{alpasim_2025}. Across the case studies, scenario features are obtained by concatenating the DINOv3 camera embedding with the ego-vehicle's body-frame \(x\)-\(y\) velocity and acceleration at the open-loop prediction timestamp or closed-loop simulation engagement timestamp. All reported results are variance reductions relative to the simple Monte Carlo estimator.

\begin{figure*}
    \centering
    \includegraphics[width=0.8\columnwidth]{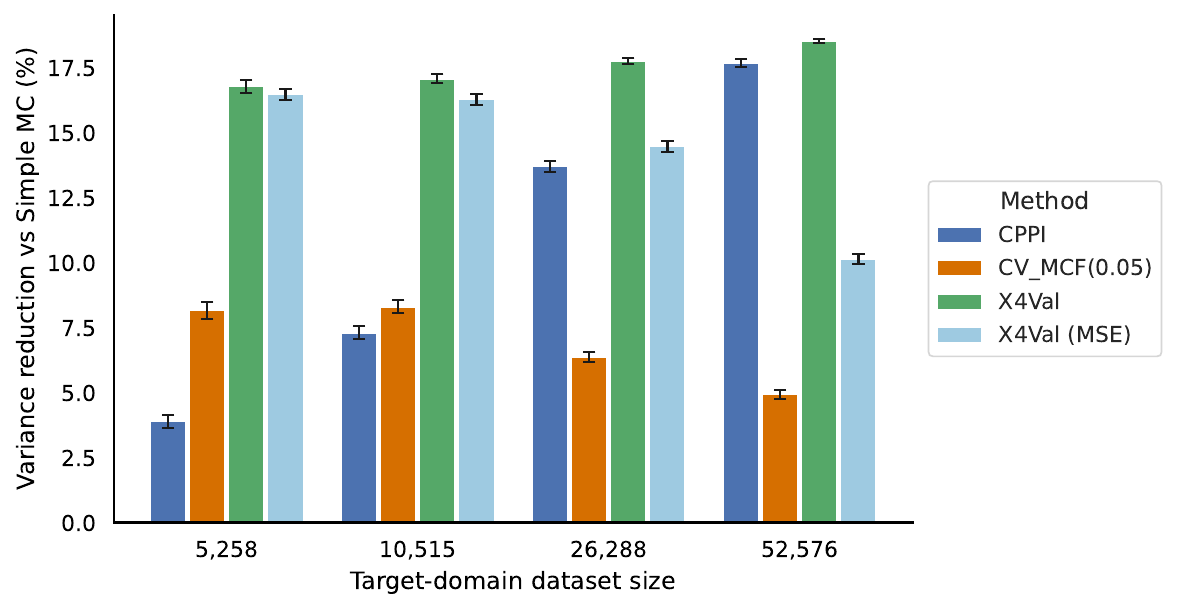}
    \caption{AV deployment to a new region. With limited target-domain evaluation data from a new region (Germany), leveraging auxiliary data from past evaluations in another region (United States) can reduce variance of performance estimation in the target region. X4Val most efficiently combines data from tests in Germany with auxiliary data to yield consistent variance reduction relative to baselines. Error bars show 95\% standard error over 30 random seeds.}
    \label{fig:case_study_2_variance_comps}
\end{figure*}

\begin{figure}[t]
\centering
\begin{subfigure}{0.48\linewidth}
    \centering
    \includegraphics[width=\linewidth]{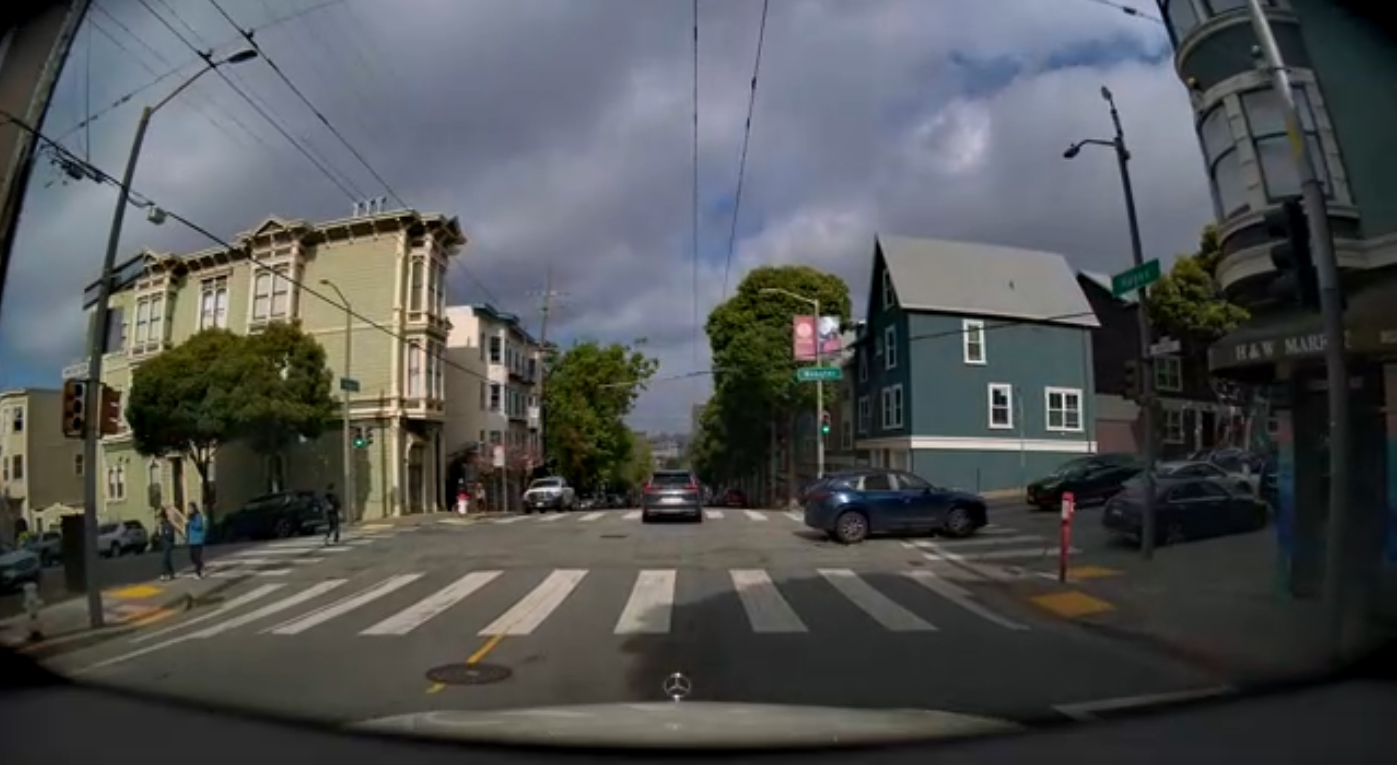}
    \caption{US driving example}
    \label{fig:us_wfov}
\end{subfigure}
\hfill
\begin{subfigure}{0.48\linewidth}
    \centering
    \includegraphics[width=\linewidth]{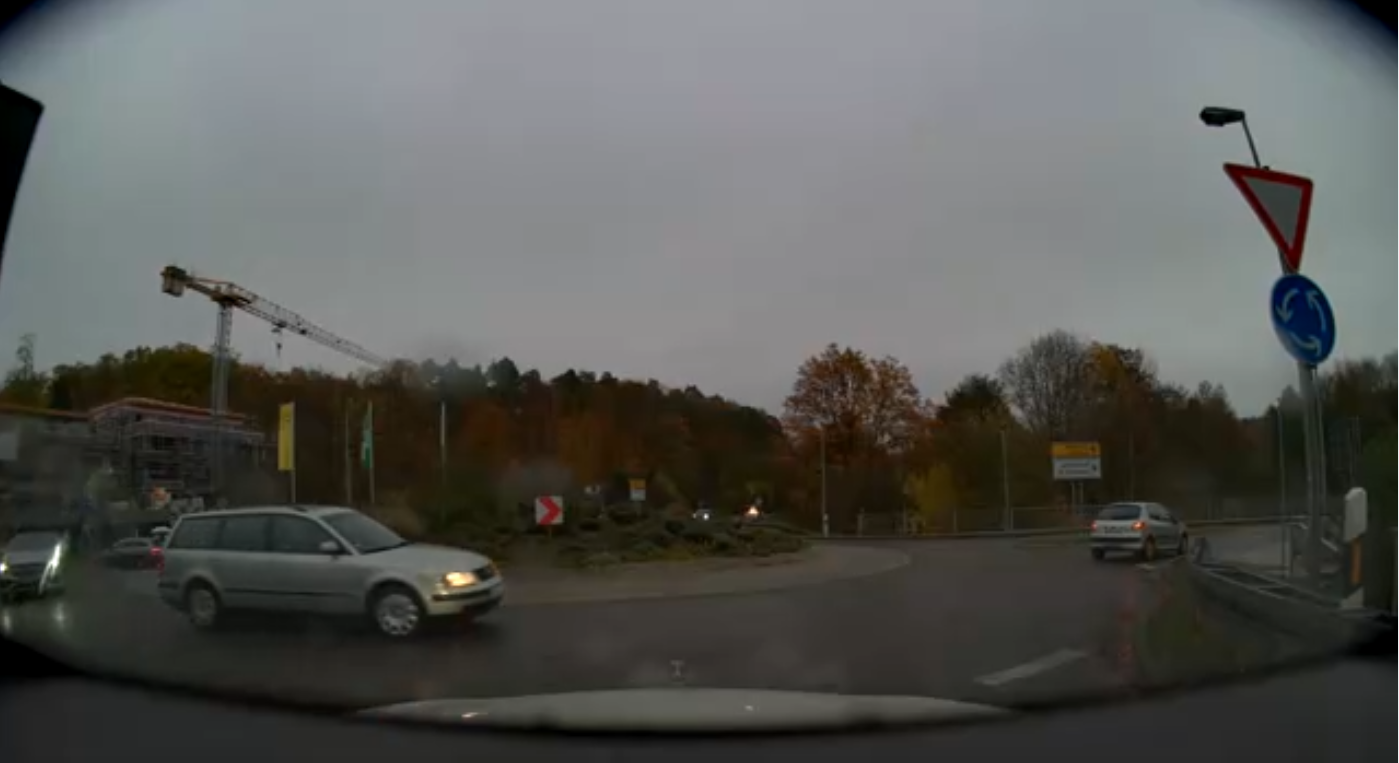}
    \caption{Germany driving example}
    \label{fig:germany_wfov}
\end{subfigure}
\caption{Driving examples for the US and Germany geographical regions. Differences include lane-marker types, signage, architecture, typical road features/junctions, etc.}
\label{fig:driving_us_germany_wfov}
\end{figure}

\subsection{Case Study 1: Autonomous Vehicle Deployment to a New Geographic Region}
In this case study, we consider a scenario in which an AV policy has been trained and validated in one geographic region, and must now be validated for deployment in a new region. We simulate this setup by training the policy on clips collected in the United States, and treating Germany as the target deployment domain. We assume that the policy has already been evaluated on held-out US clips, yielding source-region auxiliary data. To simulate validation in the new region, we collect target-domain metric measurements on clips from Germany. We further assume access to additional target-domain scenario-only data from Germany, such as data that might be obtained through a human-driven data collection campaign. We simulate this by using 86{,}848 scenario-feature samples from distinct, identically distributed clips collected in Germany. Example driving scenes from the two regions are shown in Fig.~\ref{fig:driving_us_germany_wfov}.

We consider final displacement error (FDE) at 3 seconds as an example metric of interest, and evaluate a range of methods for estimating the mean FDE of the policy in Germany. The standard Monte Carlo (\textbf{MC}) estimator uses only the evaluation results from Germany to estimate mean performance.
\textbf{CPPI} leverages the additional scenario-only Germany data to reduce variance by using the Germany evaluation data to train a predictor for FDE, and applying this predictor on the scenario-only data.
We explore two strategies for leveraging additional auxiliary data from past tests in the United States. \textbf{CV\_MCF} is a control-variate baseline with a learned metric correlator function~\cite{luo2025_sim2val}, which we refer to here as a neural surrogate.
It first pretrains the surrogate on $(\scenario,\metric)$ pairs from the auxiliary US data, before fine-tuning on a fraction (here, $0.05$) of the available Germany evaluation data to adapt to any distribution shift. The resulting predictor is used to obtain a control variate signal on the remaining Germany evaluation data, and Germany scenario-only data. \textbf{\name} uses the same auxiliary and target-domain data, but applies the cross-fitting strategy described in Section \ref{sec:method}, so that all target-domain samples can contribute to improving the neural surrogate while still being used for estimation. We also include \textbf{X4Val (MSE)}, an ablation that keeps the same data, architecture, and cross-fitting pipeline as \name{}, but replaces the variance-aligned surrogate objective with a standard mean-squared-error objective.
Figure \ref{fig:case_study_2_variance_comps} shows results across different amounts of available Germany evaluation data. \name{} consistently provides the strongest variance reduction of between 15-20\% (i.e, it provides the tightest confidence intervals).

\subsection{Case Study 2: Iterative Policy Development of an Autonomous Driving Policy}

\begin{figure*}
    \centering
    \includegraphics[width=0.5\columnwidth]{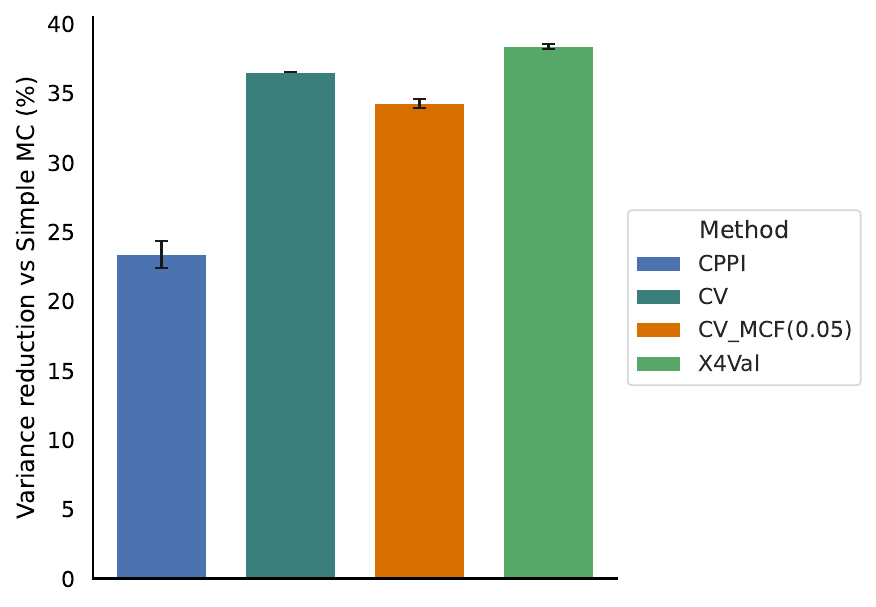}
    \caption{Iterative AV policy development. When validating a newly trained policy with limited evaluation data, historical evaluation data from earlier policy versions can serve as auxiliary information to reduce uncertainty in estimating current performance. \name{} most efficiently combines the limited current-policy evaluation data with historical data from earlier policies, achieving the largest variance reduction among all methods. Error bars show 95\% standard error over 30 random seeds.}
    \label{fig:case_study_1_variance_comps}
\end{figure*}

In this case study, we consider a setting common in large-scale autonomy programs, where policies are iteratively improved as additional training data becomes available. Over time, multiple versions of a policy are trained and evaluated, producing a corpus of historical evaluation data. When validating a newly trained policy, it is natural to ask whether this historical data can be leveraged to reduce uncertainty in estimating the current policy’s performance.

To simulate this scenario, we train a sequence of driving policies on progressively larger subsets of the NVIDIA PhysicalAI-Autonomous-Vehicles dataset \cite{nvidia2025physical_ai_av}. This produces six policy versions in total: five earlier policies trained on increasingly large subsets of the dataset, and a final policy trained on the largest dataset subset that we treat as the target for validation. For each of the five earlier policies, we assume access to 1{,}000 paired samples consisting of scenario features, open-loop metrics, and closed-loop evaluation metrics. For the current policy, we assume access to 200 paired samples, reflecting the smaller amount of validation data collected so far for a newly trained system. The evaluation scenarios for each of the earlier policies and the current policy are all distinct. In addition, we assume access to 257 scenario-only samples collected from driving in the target environment without performing closed-loop evaluation. In this experiment, the closed-loop metric serves as a proxy for the real-world performance metric. 

The standard \textbf{MC} estimator uses only the 200 evaluation samples from the current policy to estimate mean performance. Because paired open-loop and closed-loop measurements are available in this setting, a classical \textbf{CV} estimator can also be applied (as in~\cite{luo2025_sim2val}) by constructing a control variate from the open-loop metric. We additionally consider a \textbf{CV\_MCF} baseline, which trains a metric correlator function (MCF) using a small fraction of the paired current-policy data and then uses the resulting predictions as an additional control-variate signal on the remaining paired evaluation samples and scenario-only samples. We further compare against \textbf{CPPI}, which trains a predictor using current-policy data only and applies this surrogate to the additional scenario-only samples.

To effectively leverage the historical evaluation data from earlier policies, \name adopts an amortized meta-learning strategy. We first train a neural surrogate across the five earlier policies using their paired $(\scenario,\auxsignal,\metric)$ data, learning a transferable predictor that captures relationships between scenario features, inexpensive surrogate signals, and policy performance across policy versions. This predictor is then adapted to the current policy using the cross-validation strategy described in Section~\ref{sec:method}, allowing all available current-policy metric data to contribute to both adaptation and estimation. The resulting predictor is used to construct a control variate signal on the current-policy evaluation data and scenario-only samples.

In this setting, the surrogate signals derived from scenario features are already strongly correlated with the closed-loop metric (correlation $\approx 0.83$), making classical control variates effective. Nevertheless, we find that leveraging historical data from earlier policy versions provides substantial additional gains. \name reduces the variance of the estimator by 38.4\% relative to the MC baseline, outperforming CV, CPPI, and CV\_MCF; see Fig.~\ref{fig:case_study_1_variance_comps} for a comparison of various methods on iterative policy development. These results highlight that even when high-quality paired data are available for the current policy, incorporating auxiliary data from earlier policy versions can significantly improve validation efficiency.

\section{Robot Manipulation Experiments}

\begin{figure}
    \centering
    \includegraphics[width=\linewidth]{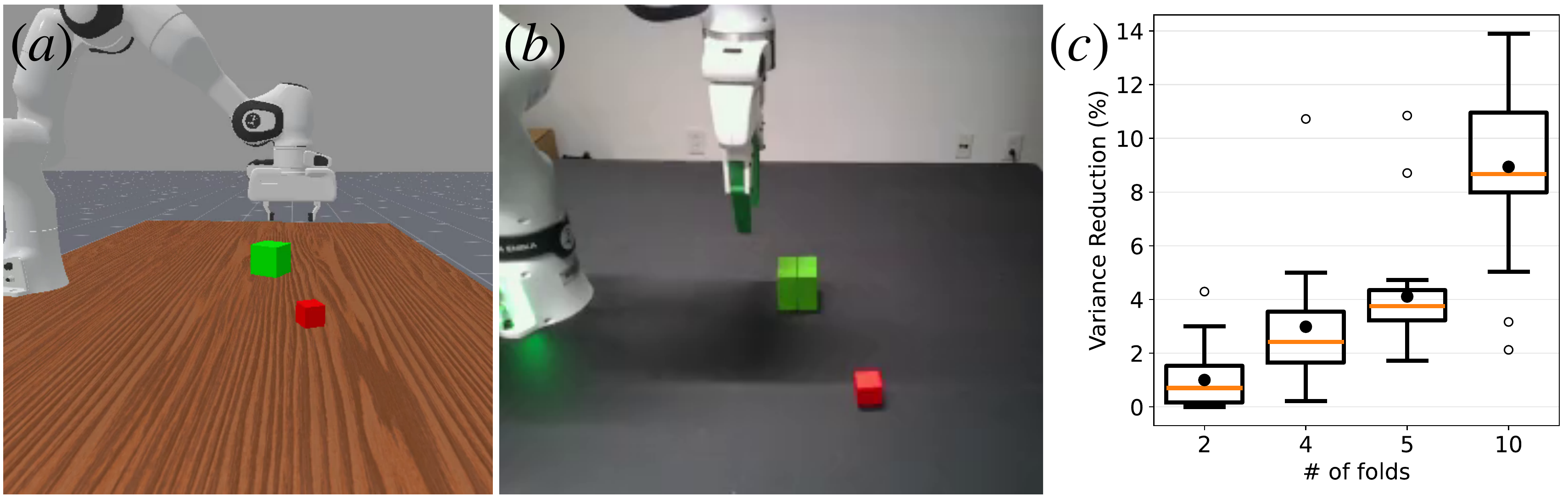}
    \vspace{-6mm}
    \caption{\name for policy evaluation in a block-stacking manipulation task.
(a) Example evaluation in the ManiSkill simulator.
(b) Example evaluation on a real robot.
(c) Variance reduction achieved by \name compared to Monte Carlo when estimating the policy’s mean success rate (each boxplot summarizes 20 random seeds).}
    \label{fig:robot-example}
    \vspace{-6mm}
\end{figure}

In this section, we demonstrate that \name enables leveraging robot manipulation policies trained in different domains to reduce the variance of policy evaluation in a target domain.

\textbf{Setup.}
We consider a block-stacking task in two environments: the ManiSkill simulator~\cite{mu2021maniskill} and a real-world setup with a Franka Panda robot arm. The goal is to train a visuomotor policy that stacks the red cube on top of the green cube, as shown in Fig.~\ref{fig:robot-example}(a)-(b). We collect 500 demonstrations in ManiSkill and 100 demonstrations in the real world, and use them to train two vision-based diffusion policies~\cite{chi2025diffusion}. We follow the default setup in~\cite{chi2025diffusion}, where each policy uses a CNN-based visual encoder that takes the two most recent images as input and predicts the next 16 actions at each timestep via a U-Net–based diffusion model.

Each policy is evaluated in its native environment. We collect 5{,}000 evaluation rollouts in ManiSkill and 100 rollouts in the real world, since real-world evaluation is significantly more time-consuming. Each rollout returns a binary success metric: 0 for failure and 1 for success. Our goal is to estimate the mean success rate of the real-world policy (which has limited evaluation data) with lower variance by leveraging evaluations of the ManiSkill policy (which has abundant data). The key intuition is that, for the block-stacking task, the relationship between policy performance and the initial block configuration may generalize across environments. For example, when the two cubes are far apart and both are far from the image center, successful task completion is difficult in both environments.

It is challenging, if not impossible, to create paired simulation and real-world rollouts as in prior work~\cite{luo2025_sim2val}. This is because the exact poses of the red and green blocks in the real-world setup, relative to the base of the Franka robot arm, are not available unless specialized infrastructure such as an indoor motion tracking system is used.

\textbf{Method.}
To apply \name to estimate the mean success rate of the real-world policy, we assume that in addition to the rollouts already collected, we can freely arrange the blocks in the real-world setup and collect images of the scene without executing the policy. In other words, we can collect initial frames of the environment without performing policy evaluation. We collect 200 such initial frames.

Our goal is to learn a predictor that maps an initial frame to the binary success metric. Since we operate in a low-data regime, we first apply DINOv2~\cite{oquab2023dinov2} to each image to obtain a 384-dimensional feature vector. We then train a simple MLP with two hidden layers of dimension 128 each to predict the binary success metric.

We adopt a pretrain–finetune transfer learning strategy: the MLP is first pretrained on the 5{,}000 ManiSkill rollouts and then finetuned on the 100 real-world rollouts. To fully utilize the limited real-world data, we employ the CPPI approach described in Section~\ref{sec:method} and split the real-world dataset into $K$ folds, where $K \in \{2,4,5,10\}$ is chosen such that $K$ divides 100. For each value of $K$, we perform 20 independent experiments with different random seeds. Within each experiment, we measure the amount of variance reduction achieved by \name relative to a simple Monte Carlo (MC) baseline that estimates the mean success rate using only the real-world rollouts.

\textbf{Results.}
Fig.~\ref{fig:robot-example}(c) presents boxplots of the percentage variance reduction across all values of $K$. The results show that \name consistently reduces the variance of the mean estimator compared with the MC baseline across all choices of $K$. Moreover, as $K$ increases, the amount of variance reduction also tends to increase. This trend is expected: as $K$ increases, the amount of training data available in each fold, $(K-1)/K \times 100$, also increases. This leads to more stable predictors across folds and consequently lower estimator variance. However, increasing $K$ also increases the computational cost, since more metric predictors must be trained.

\section{Conclusion}
We present \name, a framework for variance-reduced metric estimation that realizes the latent potential of auxiliary (non-paired, multi-domain) data, enabling more sample-efficient and rigorous validation of real-world deployed systems. 
Across autonomous driving and robot manipulation case studies, \name{} reduces estimator variance and yields tighter confidence intervals than Monte Carlo and classical control-variate baselines, demonstrating that learned neural surrogates can substantially improve the sample efficiency of rigorous policy evaluation. We also note one methodological limitation: the samples used in the final estimator, including target-domain scenario-only samples and target-domain metric samples, must be drawn from the deployment distribution whose performance is being estimated. Auxiliary data from shifted or heterogeneous domains can improve the neural surrogate, but do not themselves define the target expectation. Future work could extend this framework beyond mean estimation to other validation targets, such as quantiles, tail-risk metrics, or conditional performance estimates, and could use the learned surrogate to guide active data collection toward scenarios that are most informative for sample-efficient validation.

% ---- Bibliography ----
\bibliographystyle{splncs04}
\bibliography{main}

@String(ICML  = {Int. Conf. Mach. Learn.})

@String(ICML  = {ICML})

@inproceedings{luo2025_sim2val,
    title = {Sim2Val: Leveraging Correlation Across Test Platforms for Variance-Reduced Metric Estimation},
    author = {Rachel Luo and Heng Yang and Michael Watson and Apoorva Sharma and Sushant Veer and Edward Schmerling and Marco Pavone},
    booktitle = {Proceedings of the Conference on Robot Learning (CoRL)},
    year = {2025},
}

@article{zrnic2024cppi,
  title={Cross-prediction-powered inference},
  author={Tijana Zrnic and Emmanuel J. Cand{\`e}s},
  journal={Proceedings of the National Academy of Sciences of the United States of America},
  year={2024},
  volume={121},
}

@book{owen2013monte,
   author = {Art B. Owen},
   year = 2013,
   title = {Monte Carlo theory, methods and examples},
   publisher = {\url{https://artowen.su.domains/mc/}}
}

@inproceedings{caesar2020nuscenes,
  title={nuscenes: A multimodal dataset for autonomous driving},
  author={Caesar, Holger and Bankiti, Varun and Lang, Alex H and Vora, Sourabh and Liong, Venice Erin and Xu, Qiang and Krishnan, Anush and Pan, Yu and Baldan, Giancarlo and Beijbom, Oscar},
  booktitle={Proceedings of the IEEE/CVF conference on computer vision and pattern recognition},
  pages={11621--11631},
  year={2020}
}

@article{gulino2023waymax,
  title={Waymax: An accelerated, data-driven simulator for large-scale autonomous driving research},
  author={Gulino, Cole and Fu, Justin and Luo, Wenjie and Tucker, George and Bronstein, Eli and Lu, Yiren and Harb, Jean and Pan, Xinlei and Wang, Yan and Chen, Xiangyu and others},
  journal={Advances in Neural Information Processing Systems},
  volume={36},
  pages={7730--7742},
  year={2023}
}

@article{angelopoulos2023prediction,
  title={Prediction-powered inference},
  author={Angelopoulos, Anastasios N and Bates, Stephen and Fannjiang, Clara and Jordan, Michael I and Zrnic, Tijana},
  journal={Science},
  volume={382},
  number={6671},
  pages={669--674},
  year={2023},
  publisher={American Association for the Advancement of Science}
}

@article{angelopoulos2023ppi++,
  title={Ppi++: Efficient prediction-powered inference},
  author={Angelopoulos, Anastasios N and Duchi, John C and Zrnic, Tijana},
  journal={arXiv preprint arXiv:2311.01453},
  year={2023}
}

@article{zhou2025accelerating,
  title={Accelerating Unbiased LLM Evaluation via Synthetic Feedback},
  author={Zhou, Zhaoyi and Song, Yuda and Zanette, Andrea},
  journal={arXiv preprint arXiv:2502.10563},
  year={2025}
}

@article{boyeau2024autoeval,
  title={Autoeval done right: Using synthetic data for model evaluation},
  author={Boyeau, Pierre and Angelopoulos, Anastasios N and Yosef, Nir and Malik, Jitendra and Jordan, Michael I},
  journal={arXiv preprint arXiv:2403.07008},
  year={2024}
}

@article{levine2020offline,
  title={Offline reinforcement learning: Tutorial, review, and perspectives on open problems},
  author={Levine, Sergey and Kumar, Aviral and Tucker, George and Fu, Justin},
  journal={arXiv preprint arXiv:2005.01643},
  year={2020}
}

@inproceedings{precup2000eligibility,
  title={Eligibility traces for off-policy policy evaluation.},
  author={Precup, Doina and Sutton, Richard S and Singh, Satinder},
  booktitle={ICML},
  volume={2000},
  pages={759--766},
  year={2000},
  organization={Citeseer}
}

@inproceedings{jiang2016doubly,
  title={Doubly robust off-policy value evaluation for reinforcement learning},
  author={Jiang, Nan and Li, Lihong},
  booktitle={International conference on machine learning},
  pages={652--661},
  year={2016},
  organization={PMLR}
}

@article{mandyam2025perry,
  title={Perry: Policy evaluation with confidence intervals using auxiliary data},
  author={Mandyam, Aishwarya and Meng, Jason and Gao, Ge and Sun, Jiankai and Schwager, Mac and Engelhardt, Barbara E and Brunskill, Emma},
  journal={arXiv preprint arXiv:2507.20068},
  year={2025}
}

@article{badithela2025reliable,
  title={Reliable and scalable robot policy evaluation with imperfect simulators},
  author={Badithela, Apurva and Snyder, David and Zha, Lihan and Mikhail, Joseph and O'Kelly, Matthew and Dixit, Anushri and Majumdar, Anirudha},
  journal={arXiv preprint arXiv:2510.04354},
  year={2025}
}

@article{mu2021maniskill,
  title={Maniskill: Generalizable manipulation skill benchmark with large-scale demonstrations},
  author={Mu, Tongzhou and Ling, Zhan and Xiang, Fanbo and Yang, Derek and Li, Xuanlin and Tao, Stone and Huang, Zhiao and Jia, Zhiwei and Su, Hao},
  journal={arXiv preprint arXiv:2107.14483},
  year={2021}
}

@article{chi2025diffusion,
  title={Diffusion policy: Visuomotor policy learning via action diffusion},
  author={Chi, Cheng and Xu, Zhenjia and Feng, Siyuan and Cousineau, Eric and Du, Yilun and Burchfiel, Benjamin and Tedrake, Russ and Song, Shuran},
  journal={The International Journal of Robotics Research},
  volume={44},
  number={10-11},
  pages={1684--1704},
  year={2025},
  publisher={Sage Publications Sage UK: London, England}
}

@article{oquab2023dinov2,
  title={Dinov2: Learning robust visual features without supervision},
  author={Oquab, Maxime and Darcet, Timoth{\'e}e and Moutakanni, Th{\'e}o and Vo, Huy and Szafraniec, Marc and Khalidov, Vasil and Fernandez, Pierre and Haziza, Daniel and Massa, Francisco and El-Nouby, Alaaeldin and others},
  journal={arXiv preprint arXiv:2304.07193},
  year={2023}
}

@software{alpasim_2025,
  author       = {
    NVIDIA and
    Yulong Cao and
    Riccardo de Lutio and
    Sanja Fidler and
    Guillermo Garcia Cobo and
    Zan Gojcic and
    Maximilian Igl and
    Boris Ivanovic and
    Peter Karkus and
    Janick Martinez Esturo and
    Marco Pavone and
    Aaron Smith and
    Ellie Tanimura and
    Michal Tyszkiewicz and
    Michael Watson and
    Qi Wu and
    Le Zhang
  },
  title        = {AlpaSim: A Modular, Lightweight, and Data-Driven Research Simulator for Autonomous Driving},
  year         = {2025},
  month        = {October},
  url          = {https://github.com/NVlabs/alpasim},
}

@software{nvidia2025physical_ai_av,
  author       = {
    {NVIDIA Corporation}
  },
  title        = {{PhysicalAI-Autonomous-Vehicles} Dataset},
  year         = {2025},
  month        = {October},
  url          = {https://huggingface.co/datasets/nvidia/PhysicalAI-Autonomous-Vehicles},
}

@inproceedings{Tan2018ASO,
  title={A Survey on Deep Transfer Learning},
  author={Chuanqi Tan and Fuchun Sun and Tao Kong and Wenchang Zhang and Chao Yang and Chunfang Liu},
  booktitle={International Conference on Artificial Neural Networks},
  year={2018},
  journal={arXiv preprint arXiv:1808.01974},
}

@article{Jiang2022TransferabilityID,
  title={Transferability in Deep Learning: A Survey},
  author={Junguang Jiang and Yang Shu and Jianmin Wang and Mingsheng Long},
  journal={arXiv preprint arXiv:2201.05867},
  year={2022},
}

@article{Chato2023SurveyOT,
  title={Survey of Transfer Learning Approaches in the Machine Learning of Digital Health Sensing Data},
  author={Lina Chato and Emma E. Regentova},
  journal={Journal of Personalized Medicine},
  year={2023},
  volume={13},
}

@inproceedings{Devlin2019BERTPO,
  title={BERT: Pre-training of Deep Bidirectional Transformers for Language Understanding},
  author={Jacob Devlin and Ming-Wei Chang and Kenton Lee and Kristina Toutanova},
  booktitle={North American Chapter of the Association for Computational Linguistics},
  year={2019},
}

@article{Yosinski2014HowTA,
  title={How transferable are features in deep neural networks?},
  author={Jason Yosinski and Jeff Clune and Yoshua Bengio and Hod Lipson},
  journal={ArXiv},
  year={2014},
  volume={abs/1411.1792},
}

@article{Iakovleva2020MetaLearningWS,
  title={Meta-Learning with Shared Amortized Variational Inference},
  author={Ekaterina Iakovleva and Jakob J. Verbeek and Alahari Karteek},
  journal={arXiv preprint arXiv:2008.12037},
  year={2020},
}

@inproceedings{Ravi2018AmortizedBM,
  title={Amortized Bayesian Meta-Learning},
  author={Sachin Ravi and Alex Beatson},
  booktitle={International Conference on Learning Representations},
  year={2018},
}

@article{Zhang2025AmortizedBM,
  title={Amortized Bayesian Meta-Learning for Low-Rank Adaptation of Large Language Models},
  author={Liyi Zhang and Jake Snell and Tom Griffiths},
  journal={arXiv preprint arXiv:2508.14285},
  year={2025},
}

@inproceedings{Katdare2023MarginalizedIS,
  title={Marginalized Importance Sampling for Off-Environment Policy Evaluation},
  author={Pulkit Katdare and Nan Jiang and K. Driggs-Campbell},
  booktitle={Conference on Robot Learning},
  year={2023},
}

@article{Niu2024ACS,
  title={A Comprehensive Survey of Cross-Domain Policy Transfer for Embodied Agents},
  author={Haoyi Niu and Jianming Hu and Guyue Zhou and Xianyuan Zhan},
  journal={ArXiv},
  year={2024},
  volume={abs/2402.04580},
}

@inproceedings{Zaheer2017Deep,
  title={Deep Sets},
  author={Manzil Zaheer and Satwik Kottur and Siamak Ravanbakhsh and Barnab{\'a}s P{\'o}czos and Ruslan Salakhutdinov and Alex Smola},
  year={2017},
}

\newpage
\appendix

\section{Cross-Fitted Estimator and Confidence Intervals}
\label{app:crossfit}

This section gives the full cross-fitted version of the estimator described in Section~\ref{sec:method}. Cross-fitting allows all labeled target-domain samples to be used for estimation while ensuring that each residual is evaluated using a neural surrogate trained without that sample's label.

Let
\[
    \labeleddata = \{(\scenario_i,\auxsignal_i,\metric_i)\}_{i=1}^{\numreal}
\]
denote the labeled target-domain metric data, and let
\[
    \unlabeleddata = \{(\unlabeledscenario_r,\auxsignal_r)\}_{r=1}^{\numunlabeled}
\]
denote the target-domain scenario-only data. We partition the labeled data into \(\numfolds\) disjoint folds with index sets
\[
    I_1,\ldots,I_\numfolds,
    \qquad
    \bigcup_{\foldidx=1}^{\numfolds} I_\foldidx = \{1,\ldots,\numreal\},
    \qquad
    I_\foldidx \cap I_{\ell} = \emptyset
    \quad \text{for } \foldidx \neq \ell .
\]
Let \(n_\foldidx = |I_\foldidx|\) and \(w_\foldidx = n_\foldidx/\numreal\). For each fold \(\foldidx\), we train a neural surrogate using all labeled target-domain data outside the fold, together with all auxiliary data:
\begin{equation}
    \surrogatefn^{(\foldidx)}
    = \mathcal{A}
    \left(
        \labeleddata^{-\foldidx},
        \auxdata_1,\ldots,\auxdata_\numtasks
    \right),
\end{equation}
where \(\labeleddata^{-\foldidx}\) denotes \(\labeleddata\) with the samples in fold \(I_\foldidx\) removed.

For target-domain scenario-only samples, we define the fold-averaged surrogate
\begin{equation}
    \bar f(\embedding) = \sum_{\foldidx=1}^{\numfolds} 
    w_\foldidx \surrogatefn^{(\foldidx)}(\embedding).
\end{equation}
When the folds are equal-sized, this reduces to the simple average
\[
    \bar f(\embedding) =
    \frac{1}{\numfolds}
    \sum_{\foldidx=1}^{\numfolds}
    \surrogatefn^{(\foldidx)}(\embedding).
\]

The cross-fitted \name{} estimator is
\begin{equation}
\hat{\mu}_{\method,\mathrm{cf}} = \frac{1}{\numunlabeled}
\sum_{r=1}^{\numunlabeled}
\bar f\!\left(
    \embedfn(\unlabeledscenario_r,\auxsignal_r)
\right)
+ \frac{1}{\numreal}
\sum_{\foldidx=1}^{\numfolds}
\sum_{i\in I_\foldidx}
\left[
    \metric_i
    - \surrogatefn^{(\foldidx)}
    \left(
        \embedfn(\scenario_i,\auxsignal_i)
    \right)
\right].
\label{eq:x4val-crossfit}
\end{equation}
Equivalently, for equal-sized folds, this estimator can be viewed as averaging \(\numfolds\) fold-wise estimators, where each fold contributes a surrogate expectation term computed on \(\unlabeleddata\) and a residual correction term computed on its held-out labeled samples.

\paragraph{Unbiasedness.}
Conditional on the learned fold-wise surrogates \(\{\surrogatefn^{(\foldidx)}\}_{\foldidx=1}^{\numfolds}\), and assuming that the held-out samples in each fold are drawn from the target distribution and are not used to train their corresponding surrogate, the estimator is unbiased:
\begin{align*} 
E\!\left[ \hat{\mu}_{\method,\mathrm{cf}} \,\middle|\, \{\surrogatefn^{(\ell)}\}_{\ell=1}^{\numfolds} \right]  
&= \E_{\scenario\sim\targetdist} \left[ \bar f(\embedfn(\scenario,\auxsignal)) \right] \\ 
&\qquad\qquad + \sum_{\foldidx=1}^{\numfolds} w_\foldidx \E_{\scenario\sim\targetdist} \left[ \metric - \surrogatefn^{(\foldidx)} \left( \embedfn(\scenario,\auxsignal) \right) \right] \\ 
&= \sum_{\foldidx=1}^{\numfolds} w_\foldidx \E_{\scenario\sim\targetdist} \left[ \surrogatefn^{(\foldidx)} \left( \embedfn(\scenario,\auxsignal) \right) \right] + \E_{\scenario\sim\targetdist}[\metric] \\ 
&\qquad\qquad - \sum_{\foldidx=1}^{\numfolds} w_\foldidx \E_{\scenario\sim\targetdist} \left[ \surrogatefn^{(\foldidx)} \left( \embedfn(\scenario,\auxsignal) \right) \right] \\ 
&= \E_{\scenario\sim\targetdist}[\metric] \\ 
&= \muval . 
\end{align*}
Thus, as in the single-split estimator, the neural surrogates need not be unbiased predictors of the target metric. Bias in the surrogate predictions is corrected by the held-out residual terms, provided that the residual samples are drawn from the target deployment distribution.

\paragraph{Variance decomposition.} 
Assume that the target-domain scenario-only samples and labeled target-domain samples are independent, and condition on the learned fold-wise surrogates. Define 
\[ U = \bar f(\embedfn(\scenario,\auxsignal)), \qquad \scenario\sim\targetdist, \] 
and, for fold \(\foldidx\), 
\[ \Delta^{(\foldidx)} = \metric - \surrogatefn^{(\foldidx)} \left( \embedfn(\scenario,\auxsignal) \right), \qquad \scenario\sim\targetdist . \] 
Then the conditional variance of Eq.~\ref{eq:x4val-crossfit} is 
\begin{equation} 
\begin{aligned} \Var\!\left( \hat{\mu}_{\method,\mathrm{cf}} \,\middle|\, \{\surrogatefn^{(\ell)}\}_{\ell=1}^{\numfolds} \right) &= \frac{1}{\numunlabeled} \Var_{\scenario\sim\targetdist} \left[ U \right] \\ 
&\quad+ \frac{1}{\numreal} \sum_{\foldidx=1}^{\numfolds} w_\foldidx \Var_{\scenario\sim\targetdist} \left[ \Delta^{(\foldidx)} \right]. 
\end{aligned} 
\label{eq:x4val-crossfit-variance} 
\end{equation} 
For equal-sized folds, this becomes 
\begin{equation} 
\Var\!\left( \hat{\mu}_{\method,\mathrm{cf}} \,\middle|\, \{\surrogatefn^{(\ell)}\}_{\ell=1}^{\numfolds} \right) = \frac{1}{\numunlabeled} \Var_{\scenario\sim\targetdist} \left[ U \right] + \frac{1}{\numreal\numfolds} \sum_{\foldidx=1}^{\numfolds} \Var_{\scenario\sim\targetdist} \left[ \Delta^{(\foldidx)} \right]. 
\end{equation} 
This mirrors the variance decomposition of the single-split estimator: the first term captures uncertainty in estimating the target-domain mean of the neural surrogate, while the second term captures uncertainty in estimating the residual correction. 

\paragraph{Confidence intervals.}
For implementation, define the fold-averaged surrogate predictions on target-domain scenario-only data as
\[
    U_r =
    \bar f\!\left(
        \embedfn(\unlabeledscenario_r,\auxsignal_r)
    \right),
    \qquad
    r=1,\ldots,\numunlabeled,
\]
and the held-out residuals as
\[
    \Delta_i
    = \metric_i - \surrogatefn^{(\foldidx)}
    \left(
        \embedfn(\scenario_i,\auxsignal_i)
    \right),
    \qquad
    i\in I_\foldidx .
\]
Let
\[
    \hat{\sigma}_f^2
    = \widehat{\Var}_{r=1,\ldots,\numunlabeled}
    \left[
        U_r
    \right],
    \qquad
    \hat{\sigma}_\Delta^2
    = \widehat{\Var}_{i=1,\ldots,\numreal}
    \left[
        \Delta_i
    \right].
\]
Then the cross-fitted confidence interval takes the form
\begin{equation}
\mathcal{C}^{\mathrm{cf}}_\alpha =
\left[
\hat{\mu}_{\method,\mathrm{cf}}
\pm z_{1-\alpha/2}
\sqrt{
    \frac{\hat{\sigma}_f^2}{\numunlabeled}
    + \frac{\hat{\sigma}_\Delta^2}{\numreal}
}
\right].
\label{eq:x4val-crossfit-ci}
\end{equation}
This is the cross-fitted analogue of the single-split confidence interval in Section~\ref{sec:method}. The first variance term captures uncertainty in the estimated target-domain mean of the neural surrogate, while the second captures uncertainty in the held-out residual correction. Notably, in contrast to the single-split confidence interval \eqref{eq:x4val-ci}, the residual term is divided by \(\numreal\), rather than \(\numest\), because every labeled target-domain sample contributes one held-out residual in the cross-fitted estimator.

\section{Additional Experimental Details}

The goal of the autonomous driving case studies is to estimate the expected closed-loop performance of a target autonomous driving policy. The policy is a motion planner trained via imitation learning. In Case Study 1, the target metric is the open-loop final displacement error (after 3s) in Germany, a hypothetical new deployment domain where we have performed limited testing.
In Case Study 2, the target metric is closed-loop distance to the ground truth trajectory, measured 3 seconds after the policy engagement event. Evaluating this metric requires running full closed-loop simulation, which is computationally expensive. 

\subsection{Data}

\paragraph{Scenario Features. } Each driving scenario is represented by a 768-dimensional embedding extracted from a DINOv3 ViT-B/16 model applied to the front-wide camera image at the engagement frame. We additionally include the ego-vehicle's x-y body-frame velocity and acceleration as part of the scenario features, yielding a 772-dimensional feature vector. In Case Study 2, we additionally include the open-loop final displacement error, yielding a 773-dimensional feature vector. 

\paragraph{Control Variate. } 
In Case Study 1, we assume no additional control variate signals.
In Case Study 2, all control-variate-based methods use the open-loop FDE as a control variate. This quantity can be computed without closed-loop simulation and is correlated with the target closed-loop metric.

\subsection{Methods}

\paragraph{Simple Monte Carlo. } The sample mean and sample variance are computed directly from the paired observations. No auxiliary data or control variates are used.

\paragraph{Control Variates (CV). } (Only applicable for Case Study 2) Classical control variates using open-loop FDE as the control variate. 

\paragraph{Control Variates with Metric Correlator Function (CV-MCF). } This method trains a neural network to predict the target metric of interest from scenario features, then uses its predictions as an additional control variate. In Case Study 1, this prediction is the only control variate used. In Case Study 2, this prediction is used alongside the open-loop FDE. The model architecture is a 2-hidden-layer MLP with ReLU activations, and it is trained with an AdamW optimizer with learning rate 2e-3 and weight decay 1e-2, and a batch size of 128. Pretraining was carried out for a maximum of 20 epochs, and fine-tuning was carried out for a maximum of 15 epochs, with 25\% of training data held out as a validation split to control early stopping.

\paragraph{Cross-Prediction-Powered Inference (CPPI). } This method uses 4-fold cross-validation to generate predictions on the full paired dataset, avoiding the train / estimation split. It uses the same MLP architecture and hyperparameters as CV-MCF. In each fold, the model is trained on 3/4 of the paired data and predicts on the held-out 1/4. No auxiliary data is used for training.

\paragraph{X4Val (Ours). } X4Val combines cross-fitted control variates with either a pretrain-finetune transfer learning approach or an amortized meta-learning approach to leverage all five auxiliary policy datasets. The pretrain-finetune approach uses the same architecture and hyperparameters as CV-MCF. The meta-learning approach uses 5 auxiliary datasets and consists of: (1) A DeepSet encoder~\cite{Zaheer2017Deep} that maps a context set of (scenario, metric) pairs to a 64-dimensional policy embedding. The per-element MLP has hidden dimension 128; the output MLP maps pooled representations to the policy embedding. (2) A shared-backbone predictor that processes scenario features through shared layers, then conditions on the policy embedding via a final head to predict the metric. Optimization is done with AdamW with a learning rate of 1e-3 and weight decay of 1e-4, and a batch size of 16. 

\section{Effect of Control-Weight Optimization in Control Variates}
\label{app:cvmcf-beta-comparison}

\begin{figure}[t]
  \centering
  \begin{subfigure}{0.48\linewidth}
    \centering
    \includegraphics[width=\linewidth]{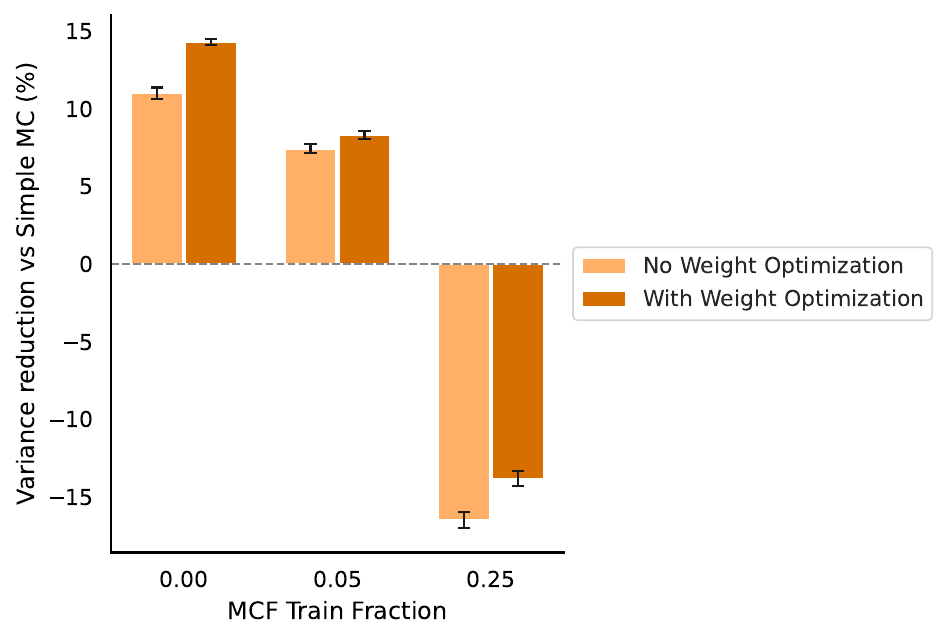}
    \caption{Case Study 1 (geographic transfer), evaluated at a target-domain dataset size of 10{,}515.}
    \label{fig:comp_optimized_beta_cs1}
  \end{subfigure}
  \hfill
  \begin{subfigure}{0.48\linewidth}
    \centering
    \includegraphics[width=\linewidth]{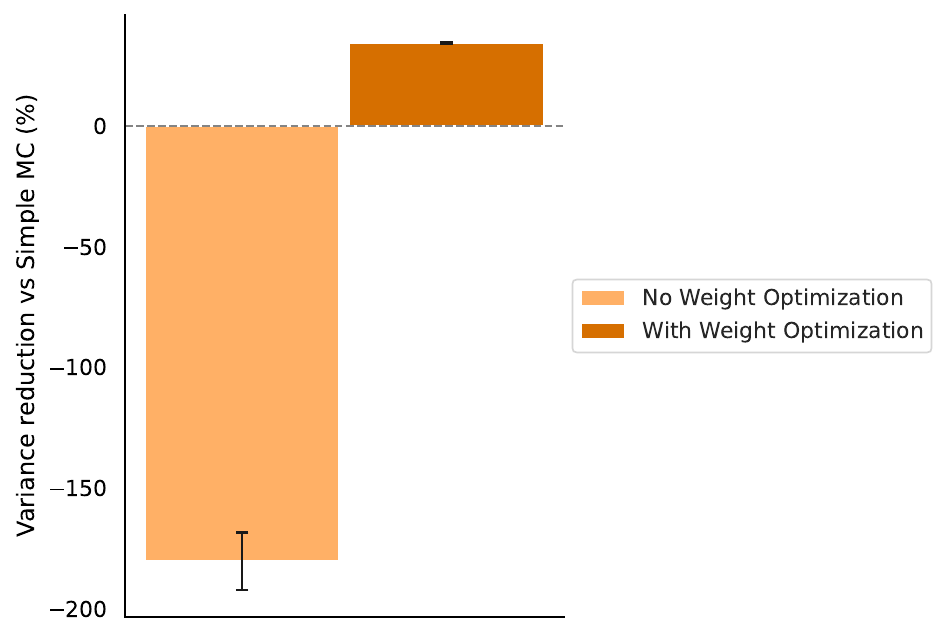}
    \caption{Case Study 2 (iterative policy development), evaluated at MCF train fraction $= 0.05$.}
    \label{fig:comp_optimized_beta_cs2}
  \end{subfigure}
  \caption{Effect of control-weight optimization on \textbf{CV\_MCF} variance reduction relative to Simple Monte Carlo. \textbf{(a)} In the geographic-transfer case study, enabling optimization yields modest but consistent gains across MCF train fractions, with the largest gain ($11\%\rightarrow 14\%$) occurring at fraction $=0.0$. \textbf{(b)} In the iterative-policy-development case study, enabling optimization is essential: it transforms an estimator that is worse than Simple MC ($-180\%$ variance reduction) into one achieving $34\%$ variance reduction. Bars show mean variance reduction across 30 random seeds; error bars are one standard error.}
  \label{fig:comp_optimized_beta}
\end{figure}

The \name{} estimator is unbiased by construction (Section~\ref{sec:learning_a_transferable_neural_surrogate}): the residual term in Eq.~\ref{eq:x4val-single-split} corrects for any systematic error in the surrogate, regardless of how the surrogate is learned. Classical control-variate methods, by contrast, typically choose the control weight $\beta$ by fitting it on the same data used to compute the residual correction, introducing a small, finite-sample bias that is usually ignored in practice. This trade-off is empirically meaningful: optimizing $\beta$ can substantially tighten the estimator's confidence intervals at the cost of introducing bias. To quantify the effect, we compare the \textbf{CV\_MCF} estimator with control-weight optimization (optimized $\beta$) and without (fixed $\beta = 1$). As shown in Fig.~\ref{fig:comp_optimized_beta}, optimization adds roughly 3 percentage points of variance reduction in Case Study 1. In Case Study 2 it is essential: the un-optimized estimator is markedly worse than Simple Monte Carlo, and enabling the optimization recovers an estimator achieving $\approx\!34\%$ variance reduction.  This asymmetry follows directly from the variance decomposition in Eq.~\ref{eq:x4val-variance}: the estimator variance is the sum of an \emph{imputed-mean} term $\widehat{\Var}(\surrogatefn)/\numunlabeled$ and a \emph{rectifier} term $\widehat{\Var}(\metric-\surrogatefn)/\numest$. Case Study 2 has only $\numunlabeled = 257$ scenario-only samples, so the imputed-mean term dominates---an un-scaled surrogate ($\beta = 1$) inflates it substantially, while optimizing $\beta$ shrinks its contribution by roughly an order of magnitude; Case Study 1, with $\numunlabeled = 86{,}848$ scenarios, leaves the imputed-mean  term negligible regardless of $\beta$.

\end{document}